\documentclass[10pt]{article}
\usepackage{amsmath,amsfonts}
\usepackage{algorithmic}
\usepackage{array}
\usepackage{graphicx}
\usepackage{textcomp}
\usepackage{hyperref}
\usepackage{booktabs}
\usepackage{tcolorbox}
\usepackage{graphicx}
\usepackage{tabularx}

\title{Evaluating Large Language Models on Computer Science University Exams in Data Structures}

\author{
Edan Gabay\thanks{These authors contributed equally to this work.},
Yael Maoz\footnotemark[1],
Jonathan Stahl\footnotemark[1],
Naama Maoz\footnotemark[1], \\
Abdo Amer\footnotemark[1],
Orr Eilat\footnotemark[1],
Hanoch Levy,
Michal Kleinbort, \\
Amir Rubinstein, 
Adi Haviv\\[1ex]
Blavatnik School of Computer Science and AI\\
Tel Aviv University
}
\date{April 28th, 2025}

\begin{document}
\maketitle
\vspace{-20pt}
\begin{abstract}
We present a comprehensive evaluation of Large Language Models (LLMs) on Computer Science (CS) Data Structure examination questions. Our work introduces a new benchmark dataset comprising exam questions from Tel Aviv University (TAU), curated to assess LLMs' abilities in handling closed and multiple-choice questions\footnote{See code at https://github.com/naamamaoz1/LLM-for-Education}.

We evaluated the performance of OpenAI's 'GPT 4o'~\cite{gpt4} and Anthropic's 'Claude 3.5'~\cite{claude}, popular LLMs, alongside two smaller LLMs, Mathstral 7B~\cite{mathstral} and LLaMA 3 8B~\cite{llama}, across the TAU exams benchmark. Our findings provide insight into the current capabilities of LLMs in CS education.

\end{abstract}

\section{Introduction}
CS education, like many other academic fields, has been profoundly impacted by students' growing access to LLMs. Students increasingly rely on these models to explain complex concepts, assist with homework assignments, debug code, and more. However, their performance in theoretical CS questions and tasks requires a thorough evaluation, due to the intricate logic and concepts of the field. 

This study aims to assess the performance of commonly used LLMs -- 'GPT 4o', 'Claude 3.5', 'Mathstral 7B' and 'LLaMA 3 8B' -- on CS theory examination questions, specifically focusing on Data Structures, which is a fundamental subject in the field. We address the following key research questions:
\begin{itemize}
    \item How accurately can LLMs solve university-level CS multiple-choice and closed-ended examination questions?
    \item How does performance vary across different topics?
    \item What is the impact of chain-of-thought (CoT) prompting strategy?
\end{itemize}

Our work introduces a novel benchmark dataset comprising of theory university-level examination questions that we translated and categorized to ensure consistent evaluation. 

Our findings indicate that 'GPT-4o' and 'Claude 3.5' achieve an overall accuracy of 56.71\% and 61.53\% on multiple-choice questions and 46.67\% and 49.33\% on closed-ended questions, significantly outperforming random guessing. However, the two smaller models achieve lower scores, with Mathstral achieving 21.95\% and LLaMA achieving 27.851\% on closed-ended questions. The models show particular strength in fundamental topics while struggling more with abstract concepts. Surprisingly, CoT prompting did not significantly improve performance.

\section{Related Work}
Recent studies have investigated the application of LLMs in educational contexts, yet their specific performance on theoretical CS questions remains understudied. 

While LLMs have demonstrated impressive capabilities in general knowledge and programming tasks \cite{llms_in_programming}, handling complex algorithmic concepts and theoretical CS questions presents unique challenges, requiring multi-step reasoning and the application of theoretical knowledge to complex scenarios.

Other related benchmarks have examined the limitations of LLMs in different domains, such as dealing with question phrasing with increased complexity~\cite{gsm_symbolic}, socially significant tasks~\cite{mmlu} and high-level multiple-choice questions in STEM fields~\cite{google_proof}. 

Our work differs from these existing benchmarks in two key aspects:
\begin{itemize}
    \item Question Diversity and Complexity: While most existing benchmarks focus on practical coding challenges or mathematical problem-solving, our benchmark specifically targets university-level theoretical questions, requiring complex reasoning.
    \item Educational Context Authenticity: Unlike other benchmarks created for other purposes, our benchmark consists of university examinations spanning multiple years, providing a more authentic assessment of LLMs' capabilities in actual educational settings.
\end{itemize}

Our benchmark provides unique insights into LLMs' performance on theoretical CS concepts in educational settings, complementing existing research on their capabilities in other domains.

\section{Methods}

\subsection{Dataset}
Our dataset comprises questions taken from 2001-2018 exams given in the Data Structure course at the School of Computer Science at Tel-Aviv university.  It includes four main questions' categories (Table \ref{tab:Question_Types}). 
\begin{table}[h]
\centering
\renewcommand{\arraystretch}{1.5} %
\setlength{\tabcolsep}{5pt} %
\begin{tabularx}{1\textwidth}{|l|l|>{\raggedright\arraybackslash}X|}
\hline
\textbf{Question Type} & \textbf{Count (final)} & \textbf{Description} \\
\hline
 A &  175 & Open questions, where the students formulate their own answers. \\
\hline
 B & 58 & Closed questions requiring short explanations, e.g., specifying an O-notation complexity bound.\\
\hline
 C & 253 & Multiple choice questions with short explanations required. \\
\hline
 D & 66 & Multiple choice questions without explanations. \\
\hline
\end{tabularx}

\caption{Question types and counts in the final dataset of TAU exams in 2001--2018}
\label{tab:Question_Types}
\end{table}
\\
{\bf Translation.} The questions and the correct final answers were translated from Hebrew to latex-formatted English using GPT-4. Questions and answers were translated separately and translations were followed by a brief manual validation.   
\\
{\bf Focus of current study.}
The current study focuses on the analysis of closed and multiple-choice questions (type B , C and D questions).

\subsection{Experimental Setting}
\subsubsection{Model}
We evaluated the responses of Anthropic Claude 3.5 and OpenAI's GPT-4o, state-of-the-art LLMs, on the chosen questions from the TAU exam dataset. Specifically, we used the models claude-3-5-sonnet-20240620, with the max tokens set to 1024 and the temperature set to 0.2 and gpt-4o-2024-08-06.

Also, we evaluated the response of Mathstral and LLaMA 3, two popular open-source models. These models are significantly smaller than GPT4 and Claude-3.5-sonnet (while the parameter count for these models is not publicly available, GPT-3 has 175B parameters \cite{gpt3}, which is at least 20 times larger). Mathstral is a model specialized in solving basic math questions \cite{mathstral}. LLaMA 3 is a state-of-the-art open-source model from Meta specialized in reasoning \cite{llama}. For the smaller models, the max token limit was set to 2048, and the temperature to 0.7.

\subsubsection{Evaluation Protocol}
Each question was queried 5 times per model, with the following considerations:
Two prompt engineering approaches were tested:
    \begin{itemize}
        \item Raw questions without modifications
        \item Chain-of-Thought (CoT) prompting
    \end{itemize}

We used a two-stage approach for each query. Initially, the model was presented with the question and instructed to solve it (see the first prompt; prompts are provided in Appendix~\ref{apx:prompts}). Subsequently, we added an additional prompt to the original query and the model's response, requesting a final answer (e.g., answer selected in multiple-choice). This sequential approach proved to be more effective than a single-prompt strategy, as it enables the model to engage in reasoning before reaching its conclusion.

\subsubsection{Performance Metrics}
\label{sec:performance_metrics}

We employed a tight accuracy metric: for a given question, the accuracy is 1 if the correct answer is the top prediction, and 0 otherwise.
We required an exact match for multiple-choice questions (type C and D). However, exact matching proved ineffective for type B questions. Alternative methods, such as verifying whether the LLM's response contained the correct answer or vice versa also yielded poor results, as equivalent answers can appear in varied forms (e.g., "O(n)" and "linear time" represent the same concept but differ in expression). Consequently, we opted to manually evaluate 50\% of the responses.

\section{Results}

\subsection{TAU Dataset}  
\subsubsection{Compiling the Full Dataset}
The raw dataset contains 883 categorized and translated questions from 57 TAU Data Structures exams (2001–2018) . Since not all had solutions, the final question-answer dataset included 552 questions (see Table \ref{tab:Question_Types}).
We randomly selected 200 (types B/C/D) for testing, each evaluated 5 times with and without Chain-of-Thought prompting.
In Table~\ref{tab:all_TAU_exams_topic} in Appendix~\ref{apx:qtopics} we provide details about the break down of the questions to topics.

\subsubsection{Success Rate for Random Guess}
\begin{figure}[h]
            \centering
            \includegraphics[width=12cm]{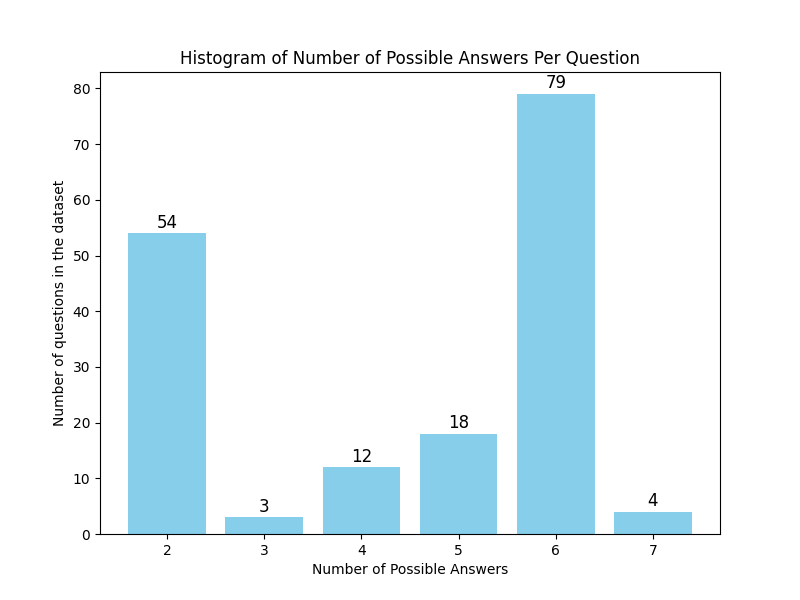}
           \caption{A histogram of the number of possible answers for each multiple choice question (C and D)}
            \label{hist_num_possible_answers}
        \end{figure}

Figure \ref{hist_num_possible_answers} depicts the distribution of the number of answers in the multiple-choice questions. 
        
        Assuming a uniform distribution, the average probability to guess the correct answer in our dataset is:
        \[
            \frac{1}{|Q|} \sum_{q \in \text{Q}} \left( \frac{1}{\text{\# possible answers for }q} \right) \ = 0.28,
        \]
    where Q is the set of questions in the dataset.

\subsection{Performance}
\subsubsection{Accuracy across Question Types}

\begin{figure}[h]
            \centering
            \includegraphics[width=12cm]{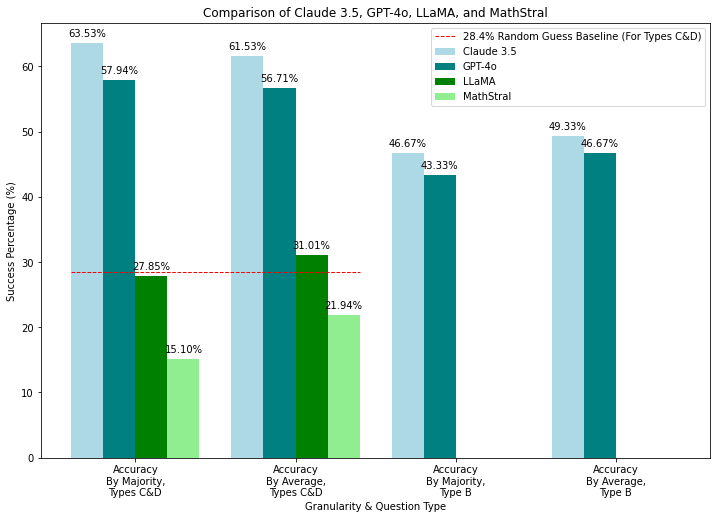}
            \caption{Accuracy by question type. Success was measured in two granularities: "Majority" (in which success for a question requires correct answers in at least 3 out of 5 repetitions per question) and "Average" (calculated across all repetitions independently). Type B success rates were calculated on the manual evaluation of 50\% of the tested questions, while Type C success rates were calculated using an exact match on the final multiple-choice answer. Red dashed line indicates the average probability to guess the correct answer in our dataset for type C and D questions. }
            \label{q_type_acc}
        \end{figure}

\begin{itemize}
        \item Type B: Overall, GPT-4o achieved a 46.67\% accuracy rate across all tested prompts, while Claude 3.5 achieved a 49.33\%. 
        
        Additionally, GPT-4o achieved a 43.33\% accuracy rate across all tested questions (a question was considered 'correctly answered' if the model provided the right answer in at least 3 out of 5 repetitions), compared to Claude 3.5's 46.67\%  (Figure \ref{q_type_acc}). Our evaluation was performed manually on 50\% of the data (see Section\ref{sec:performance_metrics}).

        For Mathstral and LLaMA 3, the success rate on type B questions was too low to be meaningful, so we excluded them.

        \item Type C and D: Overall, GPT-4o achieved a 56.71\% accuracy rate across all tested prompts for multiple-choice questions, compared to Claude 3.5's 61.53\%. 

        For the smaller models, across all tested prompts for multiple-choice questions Mathstral achieved a 21.94\% accuracy, while LLaMA achieved 31.01\%.
        
        On majority counting GPT-4o achieved a 57.94\% accuracy rate, while Claude 3.5
       achieved a 63.53\% accuracy rate (Figure \ref{q_type_acc}). 

    \end{itemize}      
       
       These results indicate that both GPT-4o and Claude 3.5 outperform random guessing, with Claude 3.5 exhibiting superior performance. However, Mathstral and LLaMA's accuracy is not significantly above the random guessing chance.

\subsubsection{Accuracy across prompt-engineering techniques} 
\begin{figure}[h]
    \centering
    \includegraphics[width=12cm]{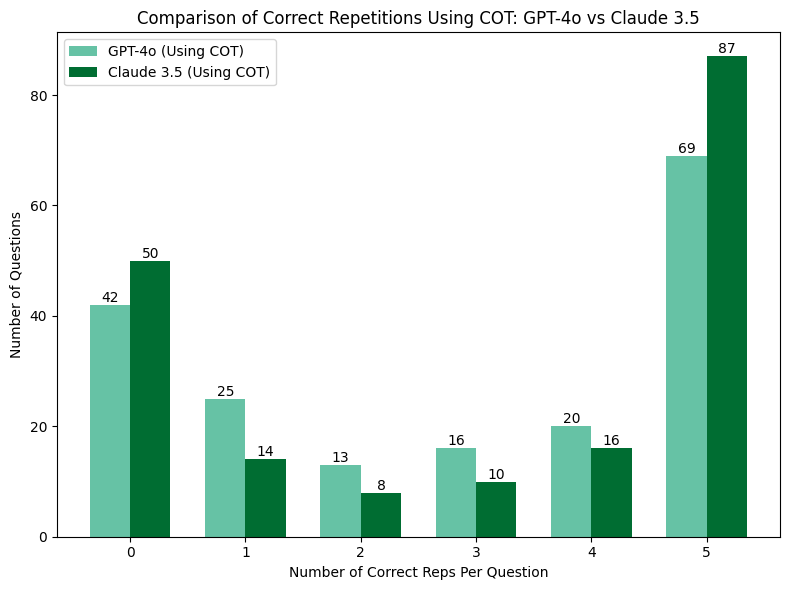}
    \caption{A histogram of the number of correct repetitions (out of 5), with CoT.}
    \label{hist_acc}
\end{figure}

We found that using a basic prompt and using chain-of-thought (CoT) prompt engineering achieved comparable results  for GPT-4o and Claude 3.5 (and the smaller LLaMA).  Figure \ref{hist_acc} depicts the results for CoT; An interesting observation is that Claude 3.5 is more "decisive" than GPT-4o, meaning that the LLM is either right or wrong across all repetitions of a question most of the time, although these results may change with different temperature configurations. 

\subsubsection{Accuracy across different question topics}
In Appendix~\ref{apx:srbytopics}, we present an analysis of the success rates of GPT and Claude across different question topics. We observe that the models perform better on some topics while performing poorly on others.

\section{Discussion}
\textbf{Accuracy by Question Type}
For all models, multiple-choice questions (Types C, D) demonstrated higher accuracy rates than Type B questions, likely due to their restricted answer space increasing probability of correct random guess.

\textbf{Accuracy by Question Topics}
Both larger models performed best on fundamental CS topics (arrays, linked lists, hash tables), possibly due to their prevalence in training data and programming practice. Conversely, both models exhibited lower performance on amortized analysis problems, potentially due to the analytical mathematical reasoning required.

\textbf{Prompt Engineering}
We assume that prompt engineering techniques have already been embedded in both models 'behind the scenes', and that this might be the reason for the poor contribution of chain-of-thought to the overall accuracy.

Our findings show that while LLMs perform significantly above chance on university-level data structures questions, their accuracy is still not sufficient for reliable academic evaluation—though they are not far from reaching that point. LLMs can be useful tools for learning and practicing, but their outputs should be approached with caution. Students must critically validate the answers provided by the model, as it is not yet fully dependable. We recommend the continued use of open-ended questions in assessments, where LLMs tend to struggle more, in order to better evaluate true student understanding. Furthermore, our findings highlight the need for using evaluation methods, specifically through supervised tests rather than homework assignments, where students cannot rely on LLMs, in order to accurately assess their independent knowledge and skills.

\bibliographystyle{IEEEtran}
\bibliography{references}

\appendix

\newpage
\section{Prompt Examples}
\label{apx:prompts}

Formats of prompts for C and D question types (the prompts for type B questions  are similar):

\begin{tcolorbox}[colback=blue!5, colframe=blue!40!black, title = First Prompt for Type C and D questions Without Prompt Engineering]
Here is a multiple-choice/true-false question from the Data Structures course. Solve the question.

\textit{[ Question Inserted Here ]}

\end{tcolorbox}

\begin{tcolorbox}[colback=blue!5, colframe=blue!40!black, title = First Prompt for Type C and D questions with CoT]

Here is a multiple-choice/true-false question from the Data Structures course. Solve the question.

Solve it step by step.

\textit{[ Question Inserted Here ]}

\end{tcolorbox}

\begin{tcolorbox}[colback=blue!5, colframe=blue!40!black, title = Second Prompt for Type C and D Questions]
\textit{[ Previous prompt and answer attached ]}

Write the final answer as a number, without additional text.
\end{tcolorbox}

\newpage
\section{Question Topics in the Dataset}
\label{apx:qtopics}

\begin{table}[h]
\centering
\begin{tabular}{|l|c|c|}
\hline
Topic Label & Topic Name & Count \\
\hline
a & Complexity analysis & 127 \\
b & Amortized Analysis & 24 \\
c & Recursive algorithms & 10 \\
d & Arrays and linked lists & 37 \\
e & Binary search trees & 50 \\
f & Balanced BST (AVL, B) & 47 \\
g & Hash tables & 47 \\
h & Binary heaps & 47 \\
i & Binomial heaps, Fibonacci heaps & 38 \\
j & Quick-sort & 11 \\
k & Lower bound comparison based sorting & 39 \\
l & Non-comparison based sorting & 8 \\
m & Selection and med-of-meds algorithm & 39 \\
n & Union-Find & 20 \\
o & Suffix trees & 7 \\
p & Other & 47 \\
\hline
\end{tabular}
\caption{Summary of question topics in the final dataset (question types B,C,D). Note that some questions were multi-topic.}
\label{tab:all_TAU_exams_topic}
\end{table}

\newpage
\section{Success Rates by Topic}
\label{apx:srbytopics}

Table~\ref{tab:cot_nocot_comparison_gpt4o_claude3} presents the success rates of GPT and Claude by the question topic.
\\
\textbf{Top Performing Topics:}  Both models performed well on arrays and linked lists (Topic d), complexity analysis (topic a), hash tables (topic g) and lower bound for comparison-based sorting (topic k). %
\\
\textbf{Low Performing Topics:} The models exhibited weak performance on the topic of selection (topic m)  and amortized analysis (topic b).

\begin{table}[h!] 
\centering 
\begin{tabular}{|c|c|c|c|c|c|} 
\hline 
\textbf{Topic} & \textbf{Q.} & \textbf{GPT} & \textbf{GPT} & \textbf{Claude} & \textbf{Claude} \\ 
& \textbf{Count} & \textbf{CoT} & \textbf{No-CoT} & \textbf{CoT} & \textbf{No-CoT} \\
\hline
a & 66 & 65\% & 63\% & 67\% & 68\% \\ b & 14 & 40\% & 39\% & 34\% & 34\% \\ c & 3 & 20\% & 13\% & 33\% & 26\% \\ d & 22 & 72\% & 65\% & 69\% & 80\% \\ e & 28 & 54\% & 52\% & 53\% & 59\% \\ f & 15 & 47\% & 47\% & 54\% & 58\% \\ g & 22 & 57\% & 58\% & 69\% & 70\% \\ h & 24 & 60\% & 62\% & 60\% & 65\% \\ i & 22 & 60\% & 56\% & 62\% & 57\% \\ j & 7 & 68\% & 53\% & 51\% & 57\% \\ k & 17 & 69\% & 51\% & 68\% & 60\% \\ l & 4 & 65\% & 70\% & 55\% & 50\% \\ m & 21 & 38\% & 49\% & 55\% & 46\% \\ n & 3 & 87\% & 100\% & 100\% & 100\% \\ o & 2 & 30\% & 40\% & 40\% & 50\% \\ p & 22 & 61\% & 59\% & 50\% & 53\% \\ 
\hline
\end{tabular} 
\caption{Comparison of success rates by topic for CoT and No-CoT methods (GPT-4o vs. Claude 3) for questions of types C and D.} 
\label{tab:cot_nocot_comparison_gpt4o_claude3} 
\end{table}

\end{document}